%% file: main.tex
\documentclass{article}
\usepackage{iclr2027_conference,times}

\usepackage[utf8]{inputenc}
\usepackage[T1]{fontenc}
\usepackage{hyperref}
\usepackage{url}
\usepackage{amsfonts}
\usepackage{nicefrac}
\usepackage{microtype}
\usepackage{xcolor}

\usepackage{tabularray}
\usepackage{tikz}
\usetikzlibrary{arrows.meta,positioning,calc,fit,backgrounds,decorations.pathreplacing}
\usepackage{pgfplots}
\pgfplotsset{compat=1.18}
\usepgfplotslibrary{groupplots,fillbetween}
\UseTblrLibrary{booktabs}
\usepackage{subcaption}
\usepackage{amsmath}
\usepackage{amssymb}
\usepackage{cleveref}
\usepackage{graphicx}
\usepackage{xspace}
\usepackage{algorithm}
\usepackage{algpseudocode}

\usepackage{wrapfig}

\newcommand{\methodname}{PrePARE\xspace}
\newcommand{\preAA}{pre-AA\xspace}

\newcommand{\postAA}{post-AA\xspace}

\newcommand{\inAA}{in-AA\xspace}
\newcommand{\best}[1]{\textbf{#1}}
\newcommand{\second}[1]{\underline{#1}}

\title{\methodname: Pre-AA Token Pruning for \\ Frozen Multi-View Geometry Transformers}

\author{
\begin{tabular*}{\dimexpr\linewidth-15pt\relax}{@{\extracolsep{\fill}}cccc@{}}
Haotang Li$^1$ & Zhenyu Qi$^1$ & Shaohan Henry Wang$^1$ & Kebin Peng$^2$ \\
Zi Wang$^3$ & Qing Guo$^4$ & Sen He$^1$ & Huanrui Yang$^1$ \\
\multicolumn{2}{c}{\normalfont $^1$University of Arizona} &
\multicolumn{2}{c}{\normalfont $^2$East Carolina University} \\
\multicolumn{2}{c}{\normalfont $^3$Augusta University} &
\multicolumn{2}{c}{\normalfont $^4$Nankai University}
\end{tabular*}
}
\iclrfinalcopy
\begin{document}

\maketitle

\input{sec/0_abstract.tex}
\input{sec/1_introduction.tex}
\input{sec/2_related.tex}
\input{sec/3_method.tex}
\input{sec/4_experiments.tex}
\input{sec/5_conclusion.tex}
\input{sec/6_statements.tex}

\bibliographystyle{iclr2027_conference}
\bibliography{references}

\appendix
\setcounter{table}{0}\renewcommand{\thetable}{A\arabic{table}}
\setcounter{figure}{0}\renewcommand{\thefigure}{A\arabic{figure}}

\end{document}

%% file: sec/0_abstract.tex
\begin{abstract}
Multi-view geometry transformers are feed-forward 3D foundation models that jointly predict depth maps, point maps, and camera poses for $N$ images in a single forward pass. Most of them build on an alternating-attention (AA) stack whose cross-view attention runs over the patch tokens of all $N$ views in every block, so its cost grows quadratically in $N$: on 300-frame ScanNetv2 clips, VGGT peaks at 72.3\,GiB, and MapAnything does not fit on a 93\,GiB card. Existing methods reduce tokens or attention inside the AA stack, so the full patch grid still enters it, and the \preAA interface, where the encoder hands its tokens to the first AA block, has been overlooked. We propose \methodname, which prunes patch tokens once at the \preAA interface and restores the dense grid once after the last AA block. Every block of the frozen AA stack then runs on the reduced tokens, while the prediction heads receive the full grid. A Token Scorer, supervised by how the unpruned AA stack uses each token, and a Feature-guided Restoration module are the only trained parts. The Token Routing module is rule-based. On ScanNetv2 at $N = 300$, \methodname reduces the peak memory of VGGT by 80\,\%, from 72.3 to 14.4\,GiB, below every \inAA method we ran, and runs 5.4 times faster, while its per-scene Chamfer distance is lower than the unpruned model's on 34 of 50 scenes (sign test $p = 0.015$). On $\pi^3$ and Depth Anything~3 it saves 15 and 20\,\% of peak memory, where the \inAA methods save at most 9\,\%, and at 1000 frames it is the only method we ran that fits Depth Anything~3 and MapAnything on one GPU.
\end{abstract}

%% file: sec/1_introduction.tex
\section{Introduction}\label{sec:intro}

Multi-view geometry transformer, e.g., VGGT \citep{VGGT}, $\pi^3$ \citep{pi3_2025}, MapAnything\citep{keetha2025mapanything}, DepthAnything3\citep{lin2025depthanything3} are feed-forward 3D foundation models that takes a set of $N$ images and jointly predicts their depth maps, point maps, and camera poses in a single forward pass.
Most of these models build on an alternating-attention (AA) stack~\citep{VGGT}. The stack alternates within-view attention, which refines the tokens of each image, with cross-view attention, which lets the tokens of all images exchange information so that the predicted depth, point maps, and camera poses agree across views. 
Because the cross-view attention runs over the patch tokens of all $N$ views in every block, its compute grows quadratically in $N$~\citep{wang2025blocksparse,ren2026speed3r}, and the AA stack becomes the main cost of the model as the input grows longer~\citep{shen2025fastvggt}. This will affect all models using AA stack in their pipeline.

\begin{figure}[t]
\centering
\vspace{-0.5cm}
\begin{subfigure}{0.54\linewidth}
    \includegraphics[width=\linewidth]{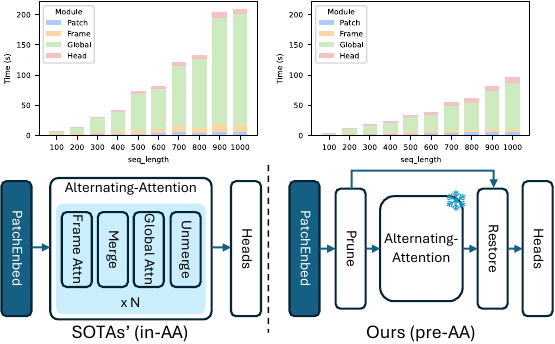}
    \caption{Structural difference between SOTAs and Ours pruning.}\label{fig:comparison}
\end{subfigure}
\hfill
\begin{subfigure}{0.45\linewidth}
    \includegraphics[width=\linewidth]{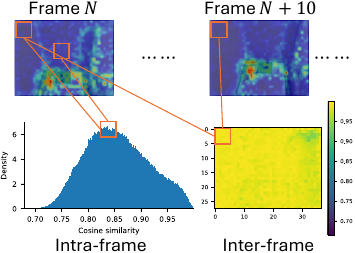}
    \caption{Pairwise cosine similarity of DINO patch tokens at the output of the frozen DINO backbone.}\label{fig:redundancy}
\end{subfigure}
\vspace{-0.6cm}
\caption{Token pruning in Geometry Transformer. (a) In-AA methods prune inside the AA stack; \methodname prunes before it and restores after it, every AA block runs on fewer tokens and latency grows slower. (b) Vision Encoder's patch tokens are highly similar within and across frames.}
\label{fig:motivation}
\vspace{-0.5cm}
\end{figure}

Recent efficient optimization works mainly focus on reducing processing in AA stack (\inAA): tokens are merged within global attention~\citep{shen2025fastvggt,shu2025litevggt,chen2025comeconfidenceguidedtokenmerging,wang2025httm}, queries and keys are reduced asymmetrically~\citep{tang2026spark3r}, or the attention operator is sparsified~\citep{wang2025blocksparse,ren2026speed3r}.
Current findings suggest that token similarity not only highly redundant among intra-frame and inter-frame (\Cref{fig:motivation}b), but also barely changes from one layer to the next~\citep{shu2025litevggt}.
Aforementioned SOTAs reduce tokens only after the full patch grid has entered the AA stack, so much of the stack still runs on all $NP$ tokens. For example, Co-Me starts merging at the 15th of VGGT's 24 blocks~\citep{chen2025comeconfidenceguidedtokenmerging}, and SAF3R makes the attention inside each block sparse~\citep{deng2026saf3r}.
A token removed at the interface of vision encoder and AA blocks (\preAA) will save the computaional cost and decision cost in every block of the stack (\Cref{fig:motivation}a), which also shows in \Cref{sec:method:prelim}, i.e., \textbf{the \preAA~pruning, potentially yield lerger efficiency improvement, has been overlooked.}

We propose \methodname (\textbf{Pre}-AA \textbf{P}runing and post-\textbf{A}A \textbf{RE}storation), which prunes patch tokens once at the \preAA interface and restores the dense patch grid once at the \postAA interface. Every block of the frozen AA stack therefore runs on the reduced tokens, and the prediction heads still receive the full grid (\Cref{fig:pipeline}).
\methodname consists of three lightweight modules.
The \textit{Token Scorer} reads each frame's encoder feature map and predicts how much the AA stack will rely on each patch token.
The \textit{Token Routing module} splits each frame's scored tokens into keep, merge, and prune sets under a per-frame keep budget. It folds each merge token into its most similar kept token in the same frame and passes the reduced sequence to the unmodified AA stack.
The \textit{Feature-guided Restoration module} rebuilds the dense grid per frame, in which the encoder features of all patch tokens query the AA outputs of the kept tokens.
The scorer has to decide before any cross-view attention runs, so we supervise it with how the unpruned AA stack uses each token. In one unpruned forward pass, we record how much attention each token receives from the anchor tokens that the pose or scale head decodes, and the largest cross-view attention it places on another view.
We first distill this AA-usage target into the scorer. We then train the scorer and the restoration module together to reproduce the unpruned model's features and predictions. The backbone stays frozen in both stages.

Because \methodname changes the backbone only at the \preAA and \postAA interfaces, we can apply the our framework to VGGT, $\pi^3$, MapAnything, and Depth Anything~3 by specifying each model's interfaces and anchor tokens (Appendix).
On ScanNetv2 at $N=300$, \methodname lowers the peak memory of VGGT by 80\,\%, from 72.3 to 14.4\,GiB, and it runs 5.4 times faster, outperforming existing SOTAs.
On MapAnything, whose unpruned model runs out of memory on 300-frame clips, \methodname runs these clips in 27.5\,GiB and 1.2 times faster than Co-Me.
On $\pi^3$ and Depth Anything~3, it saves 15 and 20\,\% of peak memory and runs 2.6 and 3.6 times faster than the respective unpruned models without a significant change in Chamfer distance, while the \inAA methods save at most 9\,\%.

The main contributions are as follows.
1) We identify the \preAA pruning's potential to optimize the latency and memory cost across the whole AA stack of a frozen multi-view geometry transformer.
2) We propose \methodname, a \preAA framework with a Token Scorer, a Token Routing module, and a Feature-guided Restoration module.
3) Evaluation on four backbone models using three datasets shows that \methodname outperforms SOTAs while maintaining the quality among all tasks. 

%% file: sec/2_related.tex
\section{Related Work}
\label{sec:related}

\noindent\textbf{Efficient geometry transformers.}
Works on making geometry transformers faster falls into four groups.
Token reduction merges or selects tokens inside the attention layers of the AA stack~\citep{shen2025fastvggt,shu2025litevggt,wang2025httm,you2026regimevggt,tang2026spark3r,zheng2026goodtokenhunting}, or merges them from the middle of the encoder onward~\citep{chen2025comeconfidenceguidedtokenmerging}.
Attention redesign makes the cross-view attention sparse~\citep{wang2025blocksparse,kim2026hess,huang2026turbovggt,ren2026speed3r,deng2026saf3r} or replaces it with cheaper operators~\citep{wang2025flashvggt,zhang2026lite3r,wang2026vggtomega}.
Sequence-level methods split long inputs into chunks that are aligned afterwards~\citep{vggtlong2025,park2026dvp,li2026svggt}, or run causally with a bounded key-value cache~\citep{streamvggt2025,infinitevggt2025,wang2026stac}.
Quantization lowers the precision of the trained model~\citep{feng2025quantvggt,pan2026qvggt}.
Most of them hand the full token grid of each frame to the first AA block.
\methodname instead prunes once at the \preAA interface and restores the full grid with a learned module, so it can run together with \inAA methods (\Cref{sec:exp:discussion}).

\noindent\textbf{Pruning criteria.}
Criteria for which tokens to keep come from feature similarity, from attention or confidence readouts, or from learned gates and rates.
ToMe~\citep{bolya2023tome} merges tokens by cosine similarity, and LiteVGGT~\citep{shu2025litevggt} adds a geometry-aware similarity.
EViT~\citep{liang2022evit} ranks tokens by the attention they receive from the class token, ATS~\citep{fayyaz2022ats} samples tokens in proportion to that attention, Co-Me~\citep{chen2025comeconfidenceguidedtokenmerging} keeps tokens by a predicted confidence, and Good Token Hunting~\citep{zheng2026goodtokenhunting} scores tokens by attention entropy.
DynamicViT~\citep{rao2021dynamicvit}, A-ViT~\citep{yin2022avit}, and IA-RED$^2$~\citep{pan2021iared2} learn per-token gates or halting, and DiffRate~\citep{chen2023diffrate} learns the compression rate of each layer.
TokenLearner~\citep{ryoo2021tokenlearner}, Evo-ViT~\citep{xu2022evovit}, and SparseViT~\citep{chen2023sparsevit} learn which tokens or windows to compute, and point-cloud transformers keep their accuracy after most of their tokens are merged~\citep{tran2025gitmerge3d}.
Each of these criteria reads the features or the attention at the place where it reduces.
The criterion of \methodname reads encoder features but learns from the cross-view attention of the unpruned AA stack, which runs after the cut.

\noindent\textbf{Trained and training-free criteria.}
Training-free methods run on released checkpoints with at most a calibration pass, and they read signals that the running model exposes, such as token similarity~\citep{shen2025fastvggt}, attention magnitude~\citep{wang2025blocksparse}, and head sensitivity~\citep{kim2026hess}.
Trained methods instead learn from targets that exist only offline.
Co-Me~\citep{chen2025comeconfidenceguidedtokenmerging} trains a confidence predictor, Speed3R~\citep{ren2026speed3r} trains a compression branch, TurboVGGT~\citep{huang2026turbovggt} learns its representative tokens, DiffRate~\citep{chen2023diffrate} learns its compression rates, and VGGT-$\Omega$~\citep{wang2026vggtomega} and Lite3R~\citep{zhang2026lite3r} train or fine-tune the network itself.
\methodname is a trained method, because its target, the attention of the unpruned AA stack on the training clips, exists only offline, and Appendix gives its training cost.

%% file: sec/3_method.tex
\section{Method}\label{sec:method}

\subsection{Preliminary}\label{sec:method:prelim}
A multi-view geometry transformer~\citep{VGGT,pi3_2025,keetha2025mapanything,lin2025depthanything3} consists of an encoder $E$, an AA stack $A$ of $L$ blocks, and prediction heads $H$.
The encoder maps each of the $N$ input frames to $P$ patch tokens with $D$ dimensions $F_f \in \mathbb{R}^{P \times D}$, and the model adds a small set $\mathcal{T}$ of special tokens, such as camera and register tokens.
The blocks of $A$ alternate within-view attention, which runs over the tokens of one frame, with cross-view attention, which runs over the tokens of all frames, $\mathcal{T} \cup \{F_f\}_{f=1}^{N}$.
The encoder is usually a separate network, such as DINOv2~\citep{oquab2023dinov2}, and in Depth Anything~3 it is the within-view prefix of a single ViT.
Dense heads read each frame's patch tokens at the last block, or at several tapped blocks for DPT-style heads~\citep{ranftl2021dpt}, and heads read a special token or pool the patch tokens.

\begin{wraptable}{r}{0.4\linewidth}
\vspace{-0.45cm}
\centering
\caption{Reduction overhead on VGGT.}

\label{tab:prelim}
\scriptsize
\begin{tblr}{
    width = \linewidth,
    colspec = {X[1.25,l]X[0.95,c]X[1.3,c]X[0.75,c]X[0.75,c]},
    colsep = 2pt, rowsep = 0.3pt,
    row{1,2} = {font=\bfseries},
}
\toprule
    & & & \SetCell[c=2]{c}Share (\%) & \\
\cmidrule[lr]{4-5}
Method & Decisions & Reduced in & Decide & Apply \\
\midrule
FastVGGT   & 24 & global attn.   & 6.0 & 4.0 \\
LiteVGGT   & 4  & global attn.   & 1.5 & 5.4 \\
\methodname & 1  & all \(L\) blocks & 0.4 & 5.0 \\
\bottomrule
\end{tblr}
\vspace{-0.4cm}
\end{wraptable}

Within-view attention costs $O(NP^2d)$ per layer and cross-view attention costs $O(N^2P^2d)$, where $d$ is the width of $A$, and the cross-view term grows quadratically with $N$.
The timing plots in \Cref{fig:comparison} show that global attention, the cross-view attention of VGGT, takes most of the forward time at every length from 100 to 1000 frames.
A token-reduction method adds the \textit{decision cost} about deciding which tokens to reduce, and \textit{applying cost} about the cost of prune step itself. \Cref{tab:prelim} gives the share of the forward pass that each method spends on each.
FastVGGT recomputes a bipartite token matching in each of the 24 global-attention layers of VGGT~\citep{shen2025fastvggt}, and LiteVGGT computes it at 4 of the 24 and reuses it~\citep{shu2025litevggt}, but both merge before and unmerge after every global attention, take 10\% and 6.9\% of total cost of the forward pass.
However, \methodname decides once at the \preAA interface, and its scoring and routing, which include the merge, take 5.4\% in total of the forward pass.
A reduction made once before $A$ pays each of the two costs once, and all $L$ blocks, with their attention, projections, and MLPs, run on the reduced tokens.
With a keep ratio $r$, both attention terms fall by $r^2$ and the linear layers by $r$ in every block.
The \preAA interface is therefore where a single reduction reaches every block of $A$ at the lowest total overhead in \Cref{tab:prelim}, but pruning there must meet two requirements: the dense heads still need the full patch grid, and the tokens must be chosen before any cross-view attention has run.

\begin{figure}[t]
    \centering
    \vspace{-0.5cm}
    \includegraphics[width=.9\linewidth]{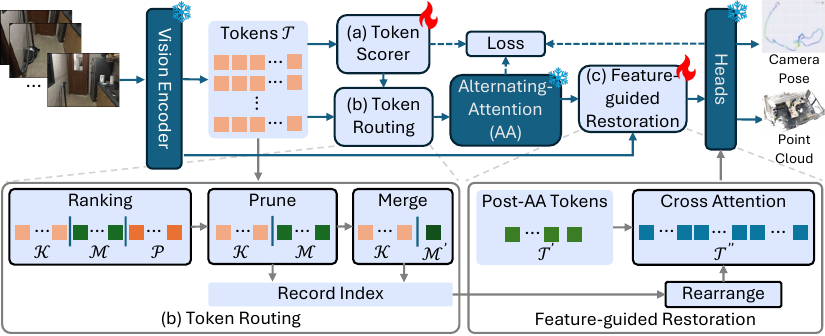}
    \vspace{-0.25cm}
    \caption{Overview of \methodname. (a) The Token Scorer scores each encoder patch token. (b) Token Routing splits the tokens into keep, merge, and prune sets \(\mathcal{K}\), \(\mathcal{M}\), \(\mathcal{P}\) under keep ratio \(r\) and merges \(\mathcal{M}\) into \(\mathcal{K}\). (c) Feature-guided Restoration rebuilds the dense grid after the frozen AA stack.}
    \label{fig:pipeline}
    \vspace{-0.7cm}
\end{figure}

\subsection{\methodname Framework}\label{sec:method:framework}
Motivatied by observation of \Cref{sec:method:prelim}, we design \methodname framework, shows in \Cref{fig:pipeline}.
The patch embedding, the AA stack, and the heads stay frozen, and only the (a) Token Scorer and the (c) Feature-guided Restoration modules are trained.
A rule-based (b) Token Routing module sits at the \preAA interface, before the first cross-view attention, and (c) Feature-guided Restoration sits at the \postAA interface, on the patch-token streams that the dense heads read.
The \preAA interface is the encoder output on multi-view geometry transformers.

The \textit{Token Scorer} reads the encoder features $F_f$ of each frame and gives every patch token a score in $[0,1]$ (\Cref{sec:method:criterion}).
\textit{Token Routing} ranks the tokens of each frame by this score, puts the top $\lceil Pr \rceil$ into the keep set $\mathcal{K}$, where $r$ is the keep ratio, and splits the rest into a merge set $\mathcal{M}$ and a prune set $\mathcal{P}$.
It drops $\mathcal{P}$, folds each token of $\mathcal{M}$ into a kept token of the same frame, and records the grid position of every kept token (\Cref{sec:method:routing}).
Scoring and routing run once for each batch before the first AA block.
The frozen AA stack then runs all $L$ blocks on the $N\lceil Pr \rceil$ kept tokens and the $|\mathcal{T}|$ special tokens, which skip routing inside AA.
\textit{Feature-guided Restoration} then uses the recorded positions to rearrange the AA outputs by frame and rebuilds the full grid of $P$ tokens per frame with cross-attention (\Cref{sec:method:routing}).
One restoration module serves each stream that a dense head reads, so each dense head receives a tensor of the same shape as in the unpruned model.
The camera head reads the camera token, which skips both routing and restoration.

The Loss block in \Cref{fig:pipeline} marks where training reads its targets, from the AA stack and from the heads (\Cref{sec:method:criterion,sec:method:training}).
The same three modules serve all four backbones, and Appendix lists what each backbone sets: the interface, the special and anchor tokens, and the tapped streams.

\subsection{Pruning Criterion}\label{sec:method:criterion}
The Token Scorer (in \Cref{fig:pipeline}-(a)) runs before any cross-view attention and sees only encoder features, so we build its criterion from the AA stack itself.
One unpruned forward pass per scene records the cross-view attention of the frozen stack, and the target $S^{\star}$ combines two readouts of this attention, one for each role a patch token plays in the stack, supporting the pose or scale estimate and matching across views.
For patch token $i$ of frame $f$, after per-frame min-max normalization $\mathrm{Norm}_f(\cdot)$, the target is
\vspace{-0.15cm}
\begin{equation}
S^{\star}(i) \;=\; \alpha \cdot \mathrm{Norm}_f\!\left(S_{\text{anchor}}(i)\right) \;+\; (1 - \alpha) \cdot \mathrm{Norm}_f\!\left(S_{\text{match}}(i)\right), \qquad \alpha \in [0,1],
\label{eq:scoregt}
\vspace{-0.2cm}
\end{equation}
where $S_{\text{anchor}}$ and $S_{\text{match}}$ are defined below and $\alpha$ sets their balance.

Let $\mathcal{A}_f$ be the anchor tokens of frame $f$, the special tokens that a pose or scale head reads. Let $\mathcal{L}_x$ be the set of cross-view blocks and $H$ the number of attention heads per block.
The anchor score averages the attention from $\mathcal{A}_f$ to $i$ over these blocks and heads,
\vspace{-0.3cm}
\begin{equation}
S_{\text{anchor}}(i) = \frac{1}{|\mathcal{L}_x|\, H\, |\mathcal{A}_f|} \sum_{l \in \mathcal{L}_x} \sum_{h=1}^{H} \sum_{a \in \mathcal{A}_f} A_{a \to i}^{l,h}, \quad
S_{\text{match}}(i) = \max_{j \neq i}\; \frac{1}{|\mathcal{L}_x|\, H} \sum_{l \in \mathcal{L}_x} \sum_{h=1}^{H} A_{i \to j}^{l,h},
\label{eq:sanchor}
\end{equation}
where $A_{u \to i}^{l,h}$ is the attention weight from query $u$ to key $i$ at head $h$ of block $l$, and $j$ runs over the other patch tokens of the scene.
A token with a high anchor score is one that the camera token of its frame attends to strongly, so the pose estimate of the unpruned model draws on it.
The matching score is high when token $i$ attends to one token elsewhere in the scene much more than to the rest, as a token does when it has a clear match in another view.
\Cref{fig:qual}a shows the tokens that $S^{\star}$ keeps.

The Token Scorer regresses $S^{\star}$ from each frame's encoder feature map $F_f \in \mathbb{R}^{h \times w \times D}$, with $hw = P$, and runs once per frame.
It compares each token with its neighbors through one self-attention layer inside $7 \times 7$ windows, and a linear layer with a sigmoid outputs $S \in [0,1]^{h \times w}$ (Appendix).
The windowed attention keeps the cost linear in $P$.
As a control, we score tokens from the encoder alone with the ToMe merge criterion~\citep{bolya2023tome} and use its complement as a distinctiveness score, $S_{\text{red}}(i) = 1 - \max_{j \in f,\, j \neq i} \cos(F_i, F_j)$, where $F_i$ and $F_j$ are the encoder feature vectors of patch tokens $i$ and $j$ in frame $f$.
\Cref{sec:exp:criterion} compares $S_{\text{red}}$ with $S^{\star}$ under the same routing and restoration.

\subsection{Routing and Restoration}\label{sec:method:routing}
A rule-based Token Routing module (in \Cref{fig:pipeline}-(b)) splits the patch tokens of each frame into a keep set $\mathcal{K}$, a merge set $\mathcal{M}$, and a prune set $\mathcal{P}$ under one keep ratio $r$ and one merge fraction $\gamma$.
The top $\lceil Pr \rceil$ tokens of each frame by score form $\mathcal{K}$, so the number of tokens that enter $A$ depends only on $r$.
The clip has a merge budget of $\gamma N P$ tokens, and each frame receives a share in proportion to the mean score of its non-kept tokens (Appendix).
Within a frame, the highest-scoring non-kept tokens fill this share and form $\mathcal{M}$, and the rest form $\mathcal{P}$.
Each merge token joins the kept token of its own frame with the most similar encoder feature, and the kept token becomes a score-weighted average of itself and the tokens that join it, with its own weight fixed at 1 (Appendix).
The same-frame rule keeps the local layout that the dense heads read after restoration, and the score weights carry the gradient of the task losses to the scorer, because top-$K$ selection is not differentiable.
Routing records which kept token each merge token joined.

After $A$, Feature-guided Restoration (in \Cref{fig:pipeline}-(c)) rebuilds the full grid of $P$ tokens of each frame at every stream $t$ that a dense head reads.
It first builds a copy-back grid $B^{t,f}$ from the routing record.
A kept position takes its own AA output, a merged position takes the AA output of the kept token it joined, and a pruned position takes an inverse-distance average of the four nearest kept positions in the grid.
A cross-attention layer then corrects this grid,
\vspace{-0.2cm}
\begin{equation}
G^{t,f}_{\text{dense}} = B^{t,f} + \lambda_{t,c} \, \mathrm{CrossAttn}\!\left(W_Q F^{f},\; W_K F^{f}_{\mathcal{K}},\; W^{t}_V G^{t,f}\right),
\label{eq:restore}
\vspace{-0.2cm}
\end{equation}
where $F^{f}$ denotes the encoder feature map $F_f$ of all $P$ positions, $F^{f}_{\mathcal{K}}$ its kept-token subset, and $G^{t,f}$ the kept tokens' AA outputs at tapped stream $t$. The matrices $W_Q$, $W_K$, and $W_V^t$ are learned query, key, and value projections, respectively.
The encoder features therefore decide which kept tokens each position reads from.
A learned gate $\lambda_{t,c}$ for each stream and each position class $c \in \{\text{kept}, \text{merged}, \text{pruned}\}$ scales the correction, so each class can be corrected by a different amount.
Restoration runs frame by frame, so its cost grows linearly with $N$.
Heads that pool patch tokens, such as the camera head of $\pi^3$, also read the restored grids, and heads that read a special token read it directly from $A$, because special tokens skip both routing and restoration.

\subsection{Training and Inference}\label{sec:method:training}
The multi-view geometry transformer stays frozen, and only the Token Scorer and the Feature-guided Restoration module are trained.
Both modules are lightweight: together they add 5.3\,M trainable parameters on VGGT, 0.42\,\% of the frozen model, and 2.1 to 7.8\,M on the other three backbone models (Appendix).
All training signals come from after the \preAA interface, either from the AA stack of the unpruned model or from the task.
Training runs in two stages.

Stage~1 trains the Token Scorer alone to fit the AA-usage target of \Cref{eq:scoregt} with a binary cross-entropy averaged over all patch tokens, $\mathcal{L}_{\mathrm{stage1}} = \frac{1}{NP}\sum_{i=1}^{NP} \mathrm{BCE}\big(S_i, S^{\star}(i)\big)$.
Stage~2 starts from this scorer and trains it together with the restoration module under $ \mathcal{L}_{\mathrm{stage2}} = \mathcal{L}_{\mathrm{stage1}} + \mathcal{L}_{\mathrm{restore}} + \mathcal{L}_{\mathrm{downstream}}$.
We use three terms because each supervises a different part of the pipeline, with $\mathcal{L}_{\mathrm{stage1}}$ choosing which tokens reach the AA stack, $\mathcal{L}_{\mathrm{restore}}$ compares each restored stream with the same stream of the unpruned stack (Appendix), and $\mathcal{L}_{\mathrm{downstream}}$ reads the predictions of the heads.
It compares the predictions of the pruned model, such as camera poses, depth or point maps, and confidences, with those of the frozen model on the same clip, and it compares the predicted depth with ground-truth depth (Appendix).
Because top-$K$ selection is not differentiable, $\mathcal{L}_{\mathrm{restore}}$ and $\mathcal{L}_{\mathrm{downstream}}$ reach the scorer only through the merge weights of \Cref{sec:method:routing}, and $\mathcal{L}_{\mathrm{stage1}}$ keeps supervising which tokens are kept and pruned.
Running distillation first and the joint stage second gives a lower Chamfer distance than either stage alone or a single joint stage (\Cref{tab:ablation}b).

At inference, one trained model serves every keep ratio.
We train at a single $r$ and choose $r$ at inference, which changes only how many tokens routing keeps in each frame (Appendix).

%% file: sec/4_experiments.tex
\section{Experiments}\label{sec:experiments}

\subsection{Setup}\label{sec:exp:setup}
\noindent\textbf{Datasets.} We train \methodname on a 262-scene building-disjoint split of ScanNetv2~\citep{dai2017ScanNetv2} and evaluate on the ScanNetv2 at \(N \in \{100, 300, 500, 1000\}\) frames, zero-shot on strides 3 and 10 of 7-Scenes~\citep{shotton20137scenes} and NRGBD~\citep{azinovic2022nrgbd}.

\noindent\textbf{Metrics.} The main text reports one dense (3D reconstruction) and one sparse (camera pose) output of the frozen heads; depth and point tracking (VGGT only) are in Appendix. Reconstruction is measured by Chamfer distance (CD) on ScanNetv2 and by accuracy, completeness and normal consistency (Acc, Comp, NC) zero-shot, and pose by ATE, ARE, RPE-rot and RPE-trans. \Cref{tab:main} also reports time per clip (Time) and peak GPU memory (Mem). Parity with the unpruned model is judged by a two-sided exact paired sign test on the per-scene metric, reported as scenes on which \methodname is better out of the paired scenes, with its \(p\)-value.

\noindent\textbf{Backbones and Baselines.} We evaluate \methodname on four backbones containing AA: VGGT~\citep{VGGT}, $\pi^3$~\citep{pi3_2025}, MapAnything (MA)~\citep{keetha2025mapanything} and Depth Anything~3-Giant (DA3)~\citep{lin2025depthanything3}, comparing the unpruned models with \inAA methods from their released code at theirs released head-sparsity profiles.

\noindent\textbf{Hyperparameters and Hardware.} \methodname uses \(\gamma = 0.30\) and \(\alpha = 0.25\) on VGGT and DA3. \(\pi^3\) and MA have no per-frame camera token and use \(\alpha = 0\). The inference keep ratio is \(r = 0.40\) on VGGT, 0.6 on \(\pi^3\), and 0.8 on MA and DA3 (Appendix). 
Training uses AdamW at learning rates \(10^{-5}\) (scorer) and \(2\times10^{-4}\) (restoration) on 2- to 8-frame clips. All runs use one 93\,GiB H100 (Appendix).

\subsection{Quantitative Results}\label{sec:exp:quant}

\methodname uses less peak memory and lower time than \inAA methods, both were measured in \Cref{tab:main}.
It also runs VGGT at 500 and 1000 frames where the unpruned models run out of GPU memory.
Its lead in camera pose also grows with the clip length: on VGGT, the ATE gap to the best \inAA method, LiteVGGT, widens from 0.012\,m at 100 frames (0.095 against 0.107) to 0.025\,m at 300 (0.092 against 0.117), and on DA3 the gap to SAF3R widens from 100 to 500 frames.
We ascribe it to the matching score.
With more frames, each surface appears in more views, so the tokens it keeps have more partners in other views and constrain the pose more tightly.
On $\pi^3$ the error between consecutive frames of \methodname reaches that of the unpruned model at 500 frames.

The saving over the unpruned model also grows with the clip length, more from 300 to 500 frames than from 100 to 300: on $\pi^3$, \methodname runs 2.3, 2.6, and 3.3 times faster at 100, 300, and 500 frames and saves 13, 15, and 33\,\% of peak memory.
The cross-view term grows with \(N^2\) and \methodname shrinks it by \(r^2\) in every block, so the unpruned time rises faster.
The part of the peak memory that does not scale with the tokens entering \(A\), such as the model weights, is a large share at 100 frames and a small share at 500, so the smaller token count shows more at longer clips.
Beyond 300 frames the memory of \methodname on VGGT grows linearly, by about 3.2\,GiB per 100 frames, and reaches 37.1\,GiB at 1000 frames.
From 300 frames on, the time gap to Co-Me grows at a nearly constant 2.2 to 2.5\,s per 100 frames, and LiteVGGT follows neither pattern, with gaps of 2.1, 15.0, and 7.6\,s.

\input{tables/tab_recon}

\noindent\textbf{3D reconstruction.}\label{sec:exp:recon} The efficiency and performance gain of \methodname improves as the clip grows compared to other methods.
At \(N = 100\), it has the best results on VGGT and DA3, second-best behind Co-Me on \(\pi^3\) (CD+0.0272) and MA (CD+0.0255).
At \(N = \{300, 500,1000\} \), \methodname reach SOTA on all four backbones among all compared methods.

\methodname removes the prune set before the AA stack, whereas LiteVGGT and Co-Me merge tokens, so every merged token still contributes to an average.
A 100-frame clip has fewer overlapping views of each surface than a longer clip of the same scene, so a pruned token more often holds content that no kept token in another view repeats, and restoration has to fill its position from its spatial neighbors.
A longer clip shows each surface in more views, so a pruned token more often has a kept counterpart in another view that cross-view attention can read.
The CD of \methodname holds at 0.4185 and 0.4187 and that of Co-Me, falls from 0.4590 to 0.4258, while those of the unpruned model, FastVGGT, and LiteVGGT rise.

\input{tables/tab_pose}

\noindent\textbf{Camera pose.}\label{sec:exp:pose} \methodname reach the SOTA among compared methods on four backbone models \Cref{tab:pose}.
We attribute this lead to the way \methodname chooses tokens, which follows the camera by design.
Both readouts of its target come from the attention that estimates the pose.
The anchor score keeps the tokens that the camera token reads, and the matching score keeps the tokens that link views, which constrain the relative pose.
The camera token skips routing and restoration, so the pose head reads it from the AA stack as in the unpruned model.
Compared methods instead choose tokens by feature similarity or by the confidence of the dense outputs, or sparsifies attention heads, so none of them reads the attention of the camera token to decide what to reduce.
The effect is largest where the camera enters the target directly.
On VGGT and DA3, whose anchor score reads a per-frame camera token, \methodname matches or improves on the unpruned model in several settings, while on $\pi^3$ and MA, which use the matching score alone, it stays behind the unpruned model in trajectory error.

\subsection{Qualitative Results}\label{sec:exp:qual}

\begin{figure}[t]
\centering
\begin{minipage}[t]{0.495\linewidth}\centering
\includegraphics[width=\linewidth]{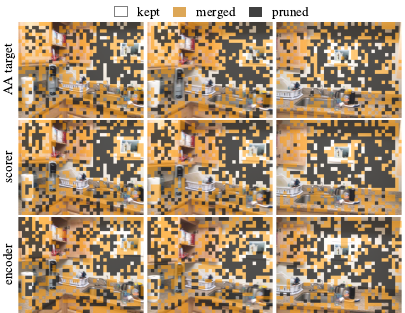}\\[-1pt]
{\small (a) Token selection at \(r = 0.4\)}
\end{minipage}\hfill
\begin{minipage}[t]{0.495\linewidth}\centering
\includegraphics[width=\linewidth]{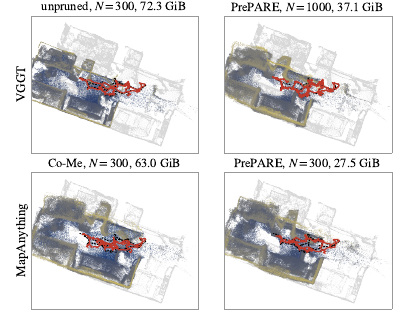}\\[-1pt]
{\small (b) Long sequences}
\end{minipage}
\vspace{-0.3cm}
\caption{\textbf{Qualitative results on ScanNetv2.} (a) Kept, merged and pruned tokens on three frames of one 4-frame clip, routed from the AA-usage target, the learned scorer and the encoder-similarity control. (b) Top-down reconstruction of one scene over the ground-truth mesh (gray), with the estimated camera path (red) over ground truth (dashed).}
\label{fig:qual}
\end{figure}

In \Cref{fig:qual}a the target and the learned scorer keep the dish rack and the picture frame and prune much of the bare wall, while the encoder control merges most of the rack; over one clip per scene, the learned keep set shares 73\,\% of the target's tokens and the control's 55\,\% (40\,\% by chance; Appendix). 
On the scene of \Cref{fig:qual}b, \methodname on VGGT at \(N = 1000\) reaches the CD of the unpruned model at \(N = 300\) (0.553) in half its memory, and on MA Co-Me reconstructs better than \methodname (0.433 against 0.497) in 2.3 times the memory.

In \Cref{fig:qual}b the estimated camera paths stay close to the ground truth in all four panels. 
On this scene ATE is 0.062\,m for the unpruned VGGT and 0.130 and 0.143\,m for \methodname and Co-Me on MA.

\subsection{Discussion}\label{sec:exp:discussion}

\input{tables/tab_discussion}

\noindent\textbf{Zero-shot transfer of the trained modules.} \methodname trains only once per backbone model, on short ScanNetv2 clips and in 2.2 to 4.0 GPU-hours (Appendix). Compared training-needed methods follows the same training protocal of \methodname.
We eveluate the ScanNetv2-trained modules on 7-Scenes and NRGBD without retraining. \Cref{tab:main_zeroshot} reports reconstruction on all 7-Scenes and NRGBD clips at strides 3 and 10 for the VGGT instance at \(r = 0.40\); against the other efficiency approaches, \methodname is more accurate and higher in NC at both strides on both datasets compared to SOTAs.

\noindent\textbf{Combining with \inAA methods.} \Cref{tab:compose} demonstrates that combining \preAA and \inAA methods without retraining lowers the cost with errors raised.
Across the four compositions, the time is 8.0 to 10.2\,s and the memory 12.1 to 16.2\,GiB, below every \inAA setting in the table, while the CD rises by 0.017 to 0.023 over the unpruned model's 0.4512 and the ATE rises from 0.096 to between 0.121 and 0.199.
The sign test against the unpruned model separates none of the composed models, the same outcome as for both \inAA methods at their milder merge, and the best composed model keeps the ATE of the \inAA methods at their default merge.
The degrades come from stacking two reductions that choose tokens in different ways.
\inAA merger works on precut token sequences and merges them by feature similarity alone, so the second cut removes content that the heads depend on, thus the pose error rises more than the CD\@.
To close the remaining gap, training the scorer and the restoration module with the merger in the stack, or letting the scorer tell the merger which tokens to keep, is a direct way to keep the lowest time and memory, which we leave to future work.

\subsection{Ablations}\label{sec:exp:ablation}

\input{tables/tab_ablation}

Lowering the keep ratio in \Cref{tab:ablation}a removes most of the cost at the first step, to $r = 0.6$, where peak memory falls from 72.3 to 17.8\,GiB and time from 48.9 to 15.8\,s, while the CD stays at parity at $r = 0.6$ and falls below the unpruned model's at $r = 0.4$, where the pruned model beats the unpruned one on 34 of 50 scenes ($p = 0.015$).
Going further to $r = 0.2$ saves little time (9.1 to 8.0\,s), most likely because the encoder, the restoration and the heads still run on the full grid, and it raises the mean CD to 0.4379 even though 31 of 50 scenes beat the unpruned model, so the extra error is concentrated in a few scenes and $r = 0.4$ stays the default.

At the same keep ratio, ToMe-style unmerging loses parity (16 of 50 scenes better, $p = 0.02$) and pushes peak memory back to 63.6\,GiB, close to the unpruned 72.3\,GiB, so the memory saving depends on how the grid is rebuilt, not only on how many tokens enter the AA stack.
Training on wide-stride clips (\textsc{ws}) keeps the peak memory of the default at 14.4\,GiB but loses parity by the same margin, so the spacing of the training frames alone can cost parity at the same token budget.

\Cref{tab:ablation}b shows that the structural choices matter more than the tuning knobs: every alternative restoration, schedule and criterion ends above the unpruned 0.4512, whereas $\gamma$ and $\alpha$ stay below it except at their extremes.
The gap that bilinear interpolation leaves to the learned restoration suggests that a pruned position needs features chosen by content, and the failing extremes suggest that merging and pruning do different jobs and that the target needs both its matching and its anchor score.

\label{sec:exp:criterion}The encoder-cosine rule reads the same encoder features as the learned scorer but gives the worst CD in panel (b), worse than zero filling, while routing by the AA-usage target $S^{\star}$ itself lands between the two, so what the scorer learns to predict matters more than what it reads.
The learned scorer also beats the target it was trained on, while stage 2 alone is the worst schedule, which suggests that the task losses can refine the selection that the target sets up but cannot find one on their own.

%% file: tables/tab_recon.tex
\begin{table}[t]
\vspace{-0.5cm}
\centering
\caption{\textbf{3D reconstruction on ScanNetv2}: Chamfer distance (CD), wall-clock per clip (Time, s) and peak memory (Mem, GiB) at \(N\) frames; first row per block: unpruned. OOM: Out of GPU Memory.}
\vspace{-0.25cm}
\label{tab:main}
\scriptsize
\begin{tblr}{
  colspec = {X[1.2,l]X[0.5,c]X[0.5,c]X[0.5,c]X[0.5,c]X[0.5,c]X[0.5,c]X[0.5,c]X[0.5,c]X[0.5,c]X[0.5,c]X[0.5,c]X[0.5,c]},
  colsep = 1.8pt, rowsep = 0.3pt,
  row{1,2} = {font=\bfseries},
  row{3,8,12,16} = {bg=gray!20},
}
\toprule
 & \SetCell[c=3]{c}100 & & & \SetCell[c=3]{c}300 & & & \SetCell[c=3]{c}500 & & & \SetCell[c=3]{c}1000 & & \\
\cmidrule[lr]{2-4}\cmidrule[lr]{5-7}\cmidrule[lr]{8-10}\cmidrule[lr]{11-13}
Method & CD\(\downarrow\) & Time\(\downarrow\) & Mem\(\downarrow\) & CD\(\downarrow\) & Time\(\downarrow\) & Mem\(\downarrow\) & CD\(\downarrow\) & Time\(\downarrow\) & Mem\(\downarrow\) & CD\(\downarrow\) & Time\(\downarrow\) & Mem\(\downarrow\) \\
\midrule
\textit{VGGT} & 0.4406 & 5.4 & 27.2 & 0.4512 & 48.9 & 72.3 & OOM & OOM & OOM & OOM & OOM & OOM \\
FastVGGT & 0.4473 & 3.0 & 8.3 & 0.4508 & 15.0 & 20.7 & 0.5014 & 35.4 & 32.5 & 0.4954 & 99.4 & 59.8 \\
Co-Me & 0.4590 & 1.9 & 9.4 & \second{0.4258} & 12.4 & 23.0 & 0.4983 & \second{29.2} & 34.7 & 0.4955 & 77.8 & 61.2 \\
LiteVGGT & \second{0.4360} & \second{1.7} & \second{7.0} & 0.4589 & \second{11.2} & \second{16.5} & \second{0.4897} & 35.9 & \second{25.8} & \second{0.4909} & \second{66.0} & \second{47.6} \\
\methodname & \best{0.4185} & \best{1.6} & \best{6.5} & \best{0.4187} & \best{9.1} & \best{14.4} & \best{0.4761} & \best{20.9} & \best{20.9} & \best{0.4883} & \best{58.4} & \best{37.1} \\
\midrule
\textit{$\pi^3$} & 0.4161 & 11.4 & 7.8 & 0.4301 & 35.9 & 15.7 & 0.4376 & 146.7 & 29.5 & 0.4306 & 365.7 & 43.9 \\
Co-Me & \best{0.3956} & 7.1 & \second{7.3} & 0.4539 & 22.2 & 15.3 & \second{0.4327} & 87.9 & \second{23.4} & \second{0.4489} & 340.9 & 43.6 \\
SAF3R & 0.4301 & \second{5.8} & 7.6 & \second{0.4320} & \second{14.9} & \second{14.3} & 0.4523 & \second{65.4} & 23.8 & 0.4849 & \second{173.5} & \second{35.9} \\
\methodname & \second{0.4228} & \best{5.0} & \best{6.8} & \best{0.4203} & \best{13.9} & \best{13.3} & \best{0.4257} & \best{45.0} & \best{19.8} & \best{0.4468} & \best{159.4} & \best{35.2} \\
\midrule
\textit{Depth Anything 3} & 0.3882 & 9.2 & 17.6 & 0.3929 & 35.4 & 40.2 & 0.4094 & 92.6 & 71.3 & OOM & OOM & OOM \\
Co-Me & \second{0.3962} & 8.5 & \second{17.0} & 0.4327 & 20.9 & \second{37.6} & \second{0.3797} & \second{98.8} & \second{58.2} & OOM & OOM & OOM \\
SAF3R & 0.3974 & \second{7.5} & 17.3 & \second{0.4187} & \second{14.3} & 43.0 & 0.4009 & 99.1 & 63.6 & OOM & OOM & OOM \\
\methodname & \best{0.3900} & \best{7.2} & \best{14.2} & \best{0.3924} & \best{9.7} & \best{32.3} & \best{0.3588} & \best{63.5} & \best{50.4} & \best{0.3721} & \best{176.2} & \best{88.9} \\
\midrule
\textit{MapAnything} & 0.3995 & 10.0 & 52.9 & OOM & OOM & OOM & OOM & OOM & OOM & OOM & OOM & OOM \\
Co-Me & \best{0.3940} & \second{7.7} & \second{36.4} & \second{0.4705} & \second{15.6} & \second{63.0} & \second{0.4820} & \second{75.8} & \second{89.7} & OOM & OOM & OOM \\
\methodname & \second{0.4195} & \best{5.1} & \best{16.0} & \best{0.4236} & \best{13.5} & \best{27.5} & \best{0.4383} & \best{63.7} & \best{42.8} & \best{0.4680} & \best{226.8} & \best{78.4} \\
\bottomrule
\end{tblr}
\vspace{-0.25cm}
\end{table}

%% file: tables/tab_pose.tex
\begin{table}[t]
\vspace{-0.1cm}
\centering
\caption{\textbf{Camera pose estimation on ScanNetv2} at \(N\) frames: ATE (m) and ARE (degrees) over the trajectory, rotation and translation RPE\textsubscript{r} (degrees) and RPE\textsubscript{t} (m) between consecutive frames, RMSE after Umeyama alignment with scale. OOM: Out of GPU Memory.}
\vspace{-0.25cm}
\label{tab:pose}
\scriptsize
\begin{tblr}{
  colspec = {X[1.1,l]X[0.5,c]X[0.5,c]X[0.5,c]X[0.5,c]X[0.5,c]X[0.5,c]X[0.5,c]X[0.5,c]X[0.5,c]X[0.5,c]X[0.5,c]X[0.5,c]},
  colsep = 2.5pt, rowsep = 0.3pt,
  row{1,2} = {font=\bfseries},
  row{3,8,12,16} = {bg=gray!20},
}
\toprule
 & \SetCell[c=4]{c}\(N{=}100\) & & & & \SetCell[c=4]{c}\(N{=}300\) & & & & \SetCell[c=4]{c}\(N{=}500\) & & & \\
\cmidrule[lr]{2-5}\cmidrule[lr]{6-9}\cmidrule[lr]{10-13}
Method & ATE\(\downarrow\) & ARE\(\downarrow\) & RPE\textsubscript{r}\(\downarrow\) & RPE\textsubscript{t}\(\downarrow\) & ATE\(\downarrow\) & ARE\(\downarrow\) & RPE\textsubscript{r}\(\downarrow\) & RPE\textsubscript{t}\(\downarrow\) & ATE\(\downarrow\) & ARE\(\downarrow\) & RPE\textsubscript{r}\(\downarrow\) & RPE\textsubscript{t}\(\downarrow\) \\
\midrule
\textit{VGGT} & 0.092 & 3.92 & 1.16 & 0.055 & 0.096 & 4.17 & 0.74 & 0.039 & OOM & OOM & OOM & OOM \\
FastVGGT & 0.118 & 4.30 & 1.71 & 0.086 & 0.123 & 4.58 & 1.13 & 0.059 & 0.129 & 4.72 & 1.05 & 0.054 \\
Co-Me & 0.132 & 4.68 & 1.94 & 0.091 & 0.127 & 4.91 & 1.36 & 0.070 & 0.134 & 4.98 & 1.24 & 0.063 \\
LiteVGGT & \second{0.107} & \second{4.06} & \second{1.42} & \second{0.072} & \second{0.117} & \second{4.46} & \second{0.95} & \second{0.048} & \second{0.122} & \second{4.63} & \second{0.91} & \second{0.046} \\
\methodname & \best{0.095} & \best{3.82} & \best{1.28} & \best{0.061} & \best{0.092} & \best{3.99} & \best{0.75} & \best{0.036} & \best{0.098} & \best{4.13} & \best{0.69} & \best{0.033} \\
\midrule
\textit{$\pi^3$} & 0.054 & 2.36 & 0.84 & 0.035 & 0.053 & 2.48 & 0.49 & 0.020 & 0.054 & 2.41 & 0.40 & 0.017 \\
Co-Me & \second{0.062} & \second{2.63} & \second{1.08} & \second{0.046} & \second{0.061} & \second{2.76} & \second{0.70} & \second{0.031} & \second{0.061} & \second{2.80} & \second{0.60} & \second{0.027} \\
SAF3R & 0.081 & 3.51 & 1.28 & 0.054 & 0.079 & 3.80 & 0.90 & 0.033 & 0.079 & 3.46 & 0.80 & 0.028 \\
\methodname & \best{0.060} & \best{2.51} & \best{0.86} & \best{0.037} & \best{0.057} & \best{2.55} & \best{0.50} & \best{0.021} & \best{0.059} & \best{2.58} & \best{0.40} & \best{0.017} \\
\midrule
\textit{Depth Anything 3} & 0.084 & 4.05 & 1.19 & 0.044 & 0.083 & 4.24 & 0.86 & 0.028 & 0.084 & 4.19 & 0.84 & 0.025 \\
Co-Me & 0.094 & 4.24 & \second{1.23} & 0.050 & 0.097 & 4.55 & 1.15 & 0.046 & 0.108 & 4.41 & 1.12 & \second{0.027} \\
SAF3R & \second{0.088} & \second{4.19} & 1.40 & \second{0.049} & \second{0.088} & \second{4.37} & \second{0.98} & \second{0.031} & \second{0.088} & \second{4.39} & \second{1.05} & \second{0.027} \\
\methodname & \best{0.086} & \best{4.05} & \best{1.19} & \best{0.046} & \best{0.085} & \best{4.32} & \best{0.92} & \best{0.029} & \best{0.081} & \best{4.20} & \best{0.92} & \best{0.026} \\
\midrule
\textit{MapAnything} & 0.127 & 5.58 & 1.73 & 0.080 & OOM & OOM & OOM & OOM & OOM & OOM & OOM & OOM \\
Co-Me & \second{0.241} & \second{11.03} & \second{2.66} & \second{0.120} & \second{0.228} & \second{11.65} & \second{1.88} & \second{0.086} & \second{0.236} & \second{11.42} & \second{1.79} & \second{0.082} \\
\methodname & \best{0.219} & \best{8.16} & \best{2.46} & \best{0.118} & \best{0.211} & \best{9.39} & \best{1.52} & \best{0.068} & \best{0.218} & \best{9.07} & \best{1.64} & \best{0.061} \\
\bottomrule
\end{tblr}
\vspace{-0.6cm}
\end{table}

%% file: tables/tab_discussion.tex
\begin{table}[t]
  \vspace{-0.3cm}
\centering
\caption{\textbf{Zero-shot 3D reconstruction}, ScanNetv2-trained modules at \(r = 0.40\).}\label{tab:main_zeroshot}
\vspace{-0.3cm}
\scriptsize
\begin{tblr}{
  colspec = {X[0.8,l]X[0.5,c]X[0.5,c]X[0.5,c]X[0.5,c]X[0.5,c]X[0.5,c]X[0.5,c]X[0.5,c]X[0.5,c]X[0.5,c]X[0.5,c]X[0.5,c]},
  colsep = 2pt, rowsep = 0.3pt,
  row{1,2} = {font=\bfseries},
  row{3} = {bg=gray!20}
}
\toprule
 & \SetCell[c=3]{c}7-Scenes Stride 3 & & & \SetCell[c=3]{c}7-Scenes Stride 10 & & & \SetCell[c=3]{c}NRGBD Stride 3 & & & \SetCell[c=3]{c}NRGBD Stride 10 & & \\
\cmidrule[lr]{2-4}\cmidrule[lr]{5-7}\cmidrule[lr]{8-10}\cmidrule[lr]{11-13}
Method & Acc\(\downarrow\) & Comp\(\downarrow\) & NC\(\uparrow\) & Acc\(\downarrow\) & Comp\(\downarrow\) & NC\(\uparrow\) & Acc\(\downarrow\) & Comp\(\downarrow\) & NC\(\uparrow\) & Acc\(\downarrow\) & Comp\(\downarrow\) & NC\(\uparrow\) \\
\midrule
\textit{VGGT} & 0.0191 & 0.0280 & 0.675 & 0.0191 & 0.0291 & 0.690 & 0.0330 & 0.0172 & 0.644 & 0.0139 & 0.0149 & 0.907 \\
Co-Me & \second{0.0206} & 0.0291 & \second{0.601} & 0.0370 & 0.0316 & \second{0.621} & \second{0.0319} & 0.0193 & 0.659 & 0.0597 & 0.0274 & \second{0.706} \\
LiteVGGT & 0.0382 & \second{0.0279} & 0.591 & \second{0.0204} & \second{0.0280} & 0.612 & 0.0329 & \second{0.0189} & \second{0.670} & \second{0.0217} & \second{0.0182} & 0.672 \\
\methodname & \best{0.0179} & \best{0.0243} & \best{0.628} & \best{0.0175} & \best{0.0276} & \best{0.651} & \best{0.0317} & \best{0.0171} & \best{0.678} & \best{0.0189} & \best{0.0161} & \best{0.777} \\
\bottomrule
\end{tblr}
\vspace{-0.7cm}
\end{table}

\begin{table}[t]
\centering
\vspace{-0.3cm}
\caption{\textbf{Composition with an \inAA merger}, ScanNetv2, \(N = 300\): sign test against the unpruned model (better scenes, \(p\)); composed rows and their matched baseline route the first frame.}\label{tab:compose}
\scriptsize
\begin{tblr}{
  colspec = {X[5,l]X[1,c]X[1,c]X[1,c]X[1,c]X[1,c]X[1,c]X[1,c]X[1.5,c]},
  colsep = 3.5pt, rowsep = 0.3pt,
  row{1} = {font=\bfseries},,
  row{2} = {bg=gray!20},
}
\toprule
Method & CD\(\downarrow\) & ATE\(\downarrow\) & ARE\(\downarrow\) & RPE\textsubscript{r}\(\downarrow\) & RPE\textsubscript{t}\(\downarrow\) & Time\(\downarrow\) & Mem\(\downarrow\) & Sign test \\
\midrule
VGGT & 0.4512 & 0.096 & 4.17 & 0.74 & 0.039 & 48.9 & 72.3 & -- \\
LiteVGGT $r{=}0.9$ & 0.4589 & 0.117 & 4.46 & 0.95 & 0.048 & 11.2 & 16.5 & 22/50, 0.48 \\
FastVGGT $r{=}0.9$ & 0.4508 & 0.123 & 4.58 & 1.13 & 0.059 & 15.0 & 20.7 & 24/50, 0.89 \\
LiteVGGT $r{=}0.6$ & 0.4258 & 0.117 & 4.11 & 0.79 & 0.034 & 10.3 & 29.9 & 31/50, 0.12 \\
FastVGGT $r{=}0.6$ & 0.4501 & 0.101 & 4.02 & 0.78 & 0.037 & 19.4 & 34.1 & 20/50, 0.20 \\
\midrule
\methodname $r{=}0.4$ + LiteVGGT $r{=}0.9$ & 0.4726 & 0.150 & 6.19& 1.54 & 0.072 & 8.0 & 12.1 & 21/50, 0.32 \\
\methodname $r{=}0.4$ + FastVGGT $r{=}0.9$ & 0.4744 & 0.199 & 7.73 & 2.38 & 0.096 & 9.4 & 14.4 & 20/50, 0.20 \\
\methodname $r{=}0.2$ + FastVGGT $r{=}0.8$ & 0.4685 & 0.137 & 5.76 & 1.41 & 0.051 & 10.2 & 16.2 & 22/50, 0.48 \\
\methodname $r{=}0.3$ + FastVGGT $r{=}0.7$ & 0.4681 & 0.121 & 5.73  & 1.31 & 0.064 & 9.7 & 15.8 & 21/50, 0.32 \\
\bottomrule
\end{tblr}
\vspace{-0.3cm}
\end{table}

%% file: tables/tab_ablation.tex
\begin{table}[t]
\centering
\caption{\textbf{Ablations} on VGGT, ScanNetv2, \(N = 300\): Time (s), Mem (GiB), sign test vs.\ unpruned (better scenes, \(p\)). Panel (b) is CD only, unpruned 0.4512 Appendix.}
\label{tab:ablation}
\scriptsize
\begin{tblr}{
  colspec = {X[2.2,l]X[0.8,c]X[0.65,c]X[0.6,c]X[0.55,c]X[1.05,c]},
  colsep = 3pt, rowsep = 0.3pt,
  row{1} = {font=\bfseries},
}
\toprule
(a) Variant & CD\(\downarrow\) & ATE\(\downarrow\) & Time & Mem & Sign test \\
\midrule
\SetCell[c=6]{l}\textit{Keep ratio ($N=300$; default scorer, learned restoration, first frame protected)} & & & & & \\
unpruned & 0.4512 & 0.096 & 48.9 & 72.3 & -- \\
$r=0.6$ & 0.4186 & 0.109 & 15.8 & 17.8 & 26/50, 0.89 \\
$r=0.4$ & 0.4187 & 0.092 & 9.1 & 14.4 & 34/50, 0.015 \\
$r=0.2$ & 0.4379 & 0.101 & 8.0 & 10.5 & 31/50, 0.12 \\
\midrule
\SetCell[c=6]{l}\textit{Restoration and training clips, against the $r=0.4$ default above} & & & & & \\
ToMe-style unmerging & 0.5229 & 0.122 & 11.2 & 63.6 & 16/50, 0.02 \\
wide-stride training clips (\textsc{ws}) & 0.4867 & 0.120 & 11.4 & 14.4 & 16/50, 0.02 \\
\end{tblr}
\begin{tblr}{
  colspec = {X[1.25,l]X[1,c]X[1,c]X[1,c]X[1,c]},
  colsep = 3pt, rowsep = 0.3pt,
  row{1} = {font=\bfseries},
}
\toprule
(b) Factor & \SetCell[c=4]{c}Variant and CD\(\downarrow\) (default in bold) & & & \\
\midrule
Restoration & zero fill 0.5136 & bilinear 0.4762 & \textbf{learned 0.4187} & \\
Merge fraction \(\gamma\) & 0: 0.4630 & \textbf{0.30: 0.4187} & 0.60: 0.4421 & 1: 0.4537 \\
Blend \(\alpha\) & 0: 0.4742 & \textbf{0.25: 0.4187} & 0.75: 0.4394 & 1: 0.4986 \\
Schedule & stage 1 only 0.4851 & stage 2 only 0.4946 & joint 0.4759 & \textbf{two-stage 0.4187} \\
Criterion & encoder cosine 0.5193 & \(S^{\star}\) oracle 0.4668 & \textbf{learned 0.4187} & \\
\bottomrule
\end{tblr}
\end{table}

%% file: sec/5_conclusion.tex
\section{Conclusion}
\label{sec:conclusion}

We presented \methodname, which prunes patch tokens once before the AA stack of a frozen multi-view geometry transformer and restores the dense grid once after it, so every AA block runs on fewer tokens. Its Token Scorer learns from how the unpruned AA stack uses each token. On ScanNetv2 at $N = 300$, it lowers the peak memory of VGGT from 72.3 to 14.4\,GiB and runs 5.4 times faster without raising the Chamfer distance. On $\pi^3$ and DA3 it saves 15 and 20\,\%, where the \inAA methods save at most 9\,\%. At 1000 frames, only \methodname fits DA3 and MapAnything on one GPU.

%% file: sec/6_statements.tex
\section*{Reproducibility Statement}

The code of this project will be released upon acceptance. The evaluation follows the FastVGGT protocol~\citep{shen2025fastvggt}: ScanNetv2~\citep{dai2017ScanNetv2} at $N \in \{100, 300, 500, 1000\}$ and, zero-shot from the ScanNetv2-trained modules, 7-Scenes~\citep{shotton20137scenes} and NRGBD~\citep{azinovic2022nrgbd} at strides 3 and 10. Metric definitions, the paired-test protocol, and the peak-memory and timing protocols are given in Appendix; the per-scene statistics behind the main-text tables are in Appendix; per-backbone interfaces, anchor tokens and loss terms are in Appendix and Appendix; training hyperparameters and cost are in Appendix. All measurements use one NVIDIA H100 NVL (93~GiB usable) in FP16 with Flash-Attention~2~\citep{dao2023flashattention2}. Training uses seed 42 with no seed replicates, evaluation point sampling seed 33, and bootstrap seed 0 with 10{,}000 resamples. The backbone models are the public checkpoints of VGGT~\citep{VGGT}, $\pi^3$~\citep{pi3_2025}, MapAnything~\citep{keetha2025mapanything} in images-only mode, and Depth Anything~3~\citep{lin2025depthanything3} (giant), and the baselines run from their released code at the settings their papers recommend.

\section*{Ethics Statement}

\methodname accelerates publicly released feed-forward 3D reconstruction models, adds 5.3~M trained parameters on VGGT (Appendix lists the other backbone models), and collects no new data. Training uses ScanNetv2 train scenes and evaluation uses ScanNetv2, 7-Scenes and NRGBD under their published terms, each of which permits non-commercial research. The backbone and baseline checkpoints carry mixed licenses (Meta's research license, BSD, Apache~2.0, MIT, and CC~BY-NC~4.0); the Co-Me and Meta terms are non-commercial, so a release of this wrapper inherits that restriction for the configurations built on them. The societal profile of the wrapped model is that of its backbone model. Lower peak memory reduces the compute and energy cost of multi-view reconstruction and allows longer clips on a single GPU, and the same capability shares the dual-use profile of the backbone models, since faster reconstruction of a scene from images can also serve surveillance or mapping of spaces without the consent of the people in them.

\section*{AI Use Statement}

Large language models were used to help with English writing and editing the manuscript, with systematic literature review before this project, and with parts of the evaluation and analysis scripts.